\documentclass{article}
\PassOptionsToPackage{numbers,sort&compress}{natbib}
\usepackage[preprint]{neurips_2026}

\title{Biologically-Inspired Spike Encoding and Optimization for Spiking Neural Networks}
\author{
  Ayaan Qayyum \\
  \texttt{aaq2109@columbia.edu} \\
  \And
  Bruce Yi Bu \\
  \texttt{yb2520@columbia.edu} \\
  \And
  Zicong Li \\
  \texttt{zl3454@columbia.edu} \\
  \And
  Jingfeng Zhou \\
  \texttt{jz3990@columbia.edu} \\
}
\usepackage{longtable}
\usepackage{amsmath, amssymb}
\usepackage{graphicx}
\usepackage{float}
\usepackage[utf8]{inputenc}
\begin{document}

\maketitle
\raggedbottom
\begin{abstract}
Spiking Neural Networks (SNNs) offer theoretical energy efficiency advantages over conventional deep networks, but to realize such advantages on vision tasks requires sparse spike representations that naive rate coding methods fail to provide. 
We combine biologically-motivated spike encoding schemes with two training improvements: Rouser adaptive threshold surrogate gradients and Kronecker-Factored Approximate Curvature (K-FAC) second-order optimization. 
The spike encoding schemes experimented with are rate coding, latency coding, DVS-style delta modulation, and Difference-of-Gaussians (DoG) filtering. 
Experiments on MNIST and Fashion-MNIST demonstrate biologically-inspired encoders provide the largest accuracy gains in capacity-constrained regimes, and is demonstrated as an effective parameter-efficiency lever. 
DoG and delta modulation encoders produce significantly sparser spike trains than rate coding. 
K-FAC yields modest but consistent accuracy improvements over Adam-trained baselines, and Rouser's learnable per-neuron thresholds improve Fashion-MNIST accuracy from 87.37\% to 88.13\% by enabling homeostatic firing-rate regulation. 
Biologically-grounded pre-processing methods can take steps to close the accuracy gap between SNNs and non-spiking baselines without increasing model size.

\end{abstract}

\section{Introduction}

Spiking Neural Networks (SNNs) are a biologically-inspired class of artificial neural networks that process information via discrete, asynchronous spike events rather than continuous activations. 
This computational paradigm mirrors the operation of biological neurons and offers potential advantages in energy efficiency since computation only occurs when a spike is generated.

However, applying SNNs to complex computer vision tasks remain a challenge. 
Standard benchmark datasets such as MNIST \cite{lecun1998mnist} already allow conventional SNNs to achieve near-ceiling accuracy
(\(\geq 98\%)\). 
In contrast, richer datasets such as Fashion-MNIST \cite{xiao2017fashion} expose weaknesses in naive rate-coded SNNs, where dense spike representations consume more energy and erode the efficiency advantage.
The central question is whether biologically-inspired pre-processing of inputs can allow smaller networks to match the performance of larger networks operating on raw pixel values. We compare three techniques: how static images are converted to temporal spike trains (spike encoding), how to backpropagate through non-differentiable spike functions (surrogate gradient learning), and whether curvature information can accelerate or improve learning (second-order optimization).

This paper makes three contributions. 
First, a controlled parameter-efficiency study of four spike encoders: rate, latency, delta modulation, and DoG, which found encoder choice may save up to 8$\times$ trainable parameters at matched accuracy for the capacity-constrained networks.  
Second, an evaluation of Rouser's adaptive threshold mechanism on the same backbone spiking MLP, showing modest yet consistent accuracy gains from essentially one extra learnable parameter per neuron. 
Third, a comparison of Adam and K-FAC as optimizers for surrogate-gradient SNN training. 
The three threads each moves SNN training toward its theoretical efficiency advantage.

\section{Background}

\subsection{Leaky Integrate-and-Fire Neurons}

The foundational building block of our SNNs is the Leaky
Integrate-and-Fire (LIF) neuron, a simplified model of biological
membrane dynamics. The subthreshold membrane potential evolves as:

\begin{equation}
u(t+1) = \beta \cdot u(t) + I(t)
\end{equation}

where \(\beta \in (0,1)\) is the membrane decay constant that controls
how quickly charge leaks away and \(I(t)\) is the synaptic input
current at time \(t\). A spike is emitted when the membrane potential
crosses a threshold \(\vartheta\):

\begin{equation}
s(t) = \Theta(u(t) - \vartheta)
\end{equation}

where \(\Theta(\cdot)\) is the Heaviside step function. After spiking,
the membrane potential is reset:

\begin{equation}
u(t) \leftarrow u(t) - \vartheta \cdot s(t)
\end{equation}

In our implementation (using snntorch), \(\vartheta = 1\) (fixed) and
\(\beta = 0.9\), with a subtractive reset mechanism.

\subsection{Problems with Gradient Descent on SNNs} \label{sec: background_gd} 

Direct training of Spiking Neural Networks (SNNs) via backpropagation is fundamentally complicated by the discontinuous, non-differentiable nature of the spike-generation function $\Theta(\cdot)$, where the true gradient is zero almost everywhere and undefined at the threshold \cite{Zenke2018}.

Extensive research has focused on bridging this gap between discrete spiking dynamics and continuous optimization frameworks.
For example, the surrogate gradient approach resolves this by substituting a smooth approximation \(\tilde{\sigma}'\) in the backward pass while retaining the hard threshold in the forward pass \cite{Neftci2019}:

\begin{equation}
\frac{\partial \mathcal{L}}{\partial u} \approx \frac{\partial \mathcal{L}}{\partial s} \cdot \tilde{\sigma}'(u - \vartheta)
\end{equation}

This effectively decouples the forward (spiking) computation from the backward (learning) computation, enabling end-to-end gradient-based training.  
However, the problem known as the 'Dead Neuron' arises because gradient flow is severed for silent neurons as $\tilde{\sigma}'(u - \vartheta) \approx 0$ when the membrane potential $u$ is consistently far below the threshold $\vartheta$.

\subsection{Spike Encoding} \label{sec: encoding}

An SNN requires a binary spike train defined across $T$ simulation timesteps rather than standard real-valued vectors as inputs.
Each pixel of an image must be converted into a spike pattern $s \in \{0,1\}^{T \times D'}$ where $D'$ is the encoder's effective channel count.
Conventional MLPs and CNNs have no analog to this conversion step.
The spike encoding process is a fixed, non-differentiable preprocessing stage sitting between the input data and the trainable section of the architecture.
It determines whether any information reaches the network at all.

Four design parameters characterize the encoder space. \textbf{Spike density} is the number of spikes per pixel per timestep. Higher density carries more bits but costs more synaptic events per inference. \textbf{Spike timing} determines when within the $T$-step window a spike arrives. A latency-coded pixel fires once, with timing as a function of pixel intensity, while a rate-coded pixel fires uniformly across time. \textbf{Channel polarity} refers to whether positive and negative deviations are carried by separate non-negative channels. This ON/OFF splitting doubles the effective input dimension. \textbf{Temporal structure} determines whether time is treated as identically-distributed sampling or as a meaningful axis carrying frame-difference or saccade-trajectory information.

These parameters are not independent. A single pixel cannot be both sparsely-encoded in time and densely-encoded in space.
Encoder choices determine the spike budget per sample, spike train dimensionality, and temporal characteristics of the spike train.

The most common choice is rate coding. Each pixel drives a Bernoulli draw, with probability equal to the normalized pixel intensity.
Rate coding is parameter-free and serves as the standard baseline in SNN literature \cite{pfeiffer2018deep}, but it is spike-inefficient.
In Section \ref{sec: results} we treat spike encoding as a meaningful optimization surface in its own right, distinct from the network-side optimizations that concern how a spike train is processed.

\subsection{Biological Inspirations in Spike Encoding}  \label{sec: bioencoding}
SNNs find their origin in biological plausibility. The majority of neurons in the brain communicate via discrete action potentials.
Many neurons also process in the continuous membrane potential domain, as seen in the largely non-spiking early vision system of invertebrates \cite{borst2023flies}.
This opens the door for encoding designs that exploit analog-domain preprocessing.

Three mechanisms from biological vision are directly relevant. The first is center-surround filtering by retinal ganglion cells (RGCs). RGC receptive fields are commonly characterized as a difference of Gaussians (DoG), where a center Gaussian responding to local light is antagonized by a wider surround Gaussian \cite{Rodieck1965}. The output is sensitive to local contrast rather than absolute intensity, suppressing spatial redundancy in natural images. The second is ON-OFF pathway separation. Bipolar cells split the center-surround signal into two non-negative streams. ON-center cells fire for bright-center/dark-surround stimuli, and OFF-center cells fire for the inverse. The two streams travel in parallel via the optic nerve and converge only in the thalamus and cortex. An equivalent ON/OFF separation appears in the fly visual system via lamina monopolar cells, in an entirely different lineage \cite{borst2023flies}. The third is microsaccadic sampling. The eye undergoes $\sim$1--2 Hz microsaccades even during fixation \cite{martinez2004role}. Microsaccades act as temporal-contrast generators on otherwise-static stimuli, driving transient ganglion cells whose responses encode change rather than persistence.

Each mechanism provides direct design inspiration for artificial spike encoders.
Center-surround filtering motivates DoG-based encoders. ON-OFF splitting motivates dual-polarity encoders. Microsaccadic motion motivates frame-difference inputs, allowing delta/DVS-style encoders to function on static images.

\subsection{An Information-Theoretic View of Spike Encoding} \label{sec: itencoding}
From an information-theoretic perspective, a spike encoder choice is essentially a different information allocation strategy.
It determines how efficiently the signal is represented in the spike domain, such as how many bits per spike the code carries, signal-to-noise ratio (SNR) and how much spatial or temporal redundancy it discards before encoding the spikes.
Since SNN energy costs scale with total spike count $\sum_t \sum_i s_t(i)$ \cite{Davies2018}, a well-designed encoder is a mechanical necessity for efficiency.
We will elaborate on each explored spike encoder's tradeoff in detail in Section \ref{sec: encoder_methodology}. 

    \subsubsection{Spatial Whitening and Mutual Information per Spike}
    Natural images have a non-uniform power spectrum $P(f) \propto 1/f^2$ \cite{Atick1992}. 
    Most of the power lives in low spatial frequencies (e.g. slowly-varying regions, homogeneous background), while high-frequency content (e.g. edges, textures) is rare and carries the bulk of the semantics.
    As such, pixel-by-pixel transmission wastes bandwidth on highly predictable signal components.
    Following the efficient-coding hypothesis \cite{barlow1961possible}, a sensory system should remove this redundancy, transmitting only the unpredictable residual.
    
    The linear filter $K(f)$ that maximizes mutual information $I(x; y)$ under a fixed output-power budget is:
    \begin{equation}
    K(f) \propto \frac{1}{\sqrt{P(f) + N(f)}}
    \end{equation}
    For realistic noise regimes, this yields a center-surround DoG kernel \cite{atick1992could}. 
    By pre-filtering with DoG, an encoder avoids wasting spikes on the predictable $1/f^2$ component, extracting more bits per spike than raw rate-coding.
    
    \subsubsection{ON/OFF Rectification and Fisher Information per Spike}
    Spike trains are inherently non-negative, thus transmitting a zero-mean signal $s$ through a single channel requires biasing or rectification, which discards information. 
    Research has shown splitting the signal into two channels doubles the Fisher information $\mathcal{F}$ per spike under a Poisson spiking model with a fixed total firing-rate budget \cite{Gjorgjieva2014}:
        \begin{equation}
        \frac{\mathcal{F}_{\text{ON/OFF}}}{\mathcal{F}_{\text{single}}} \le 2
        \end{equation}
    This $2\times$ gain is attained only when the signal is genuinely zero-mean and symmetric. 
    For strictly non-negative inputs (like raw pixels), the OFF channel is redundant and the gain disappears. 
    For a zero-mean symmetric input, the ON/OFF split therefore doubles Fisher information per spike, meaning it halves the spike count required to reach a given decoding precision.
    
    \subsubsection{Composition of Spatial Whitening and ON/OFF Rectification}
    The two aforementioned techniques complement each other by design. The whitening filter produces the zero-mean substrate required for the ON-OFF Fisher information gain.
    An encoder implementing DoG filtering followed by ON-OFF rectification is predicted to extract close to $2 \times 2 = 4\times$ as much information per spike as a standard rate code.

\subsection{Neuromorphic Hardware Considerations}

A key motivation for SNNs is their compatibility with neuromorphic hardware platforms such as Intel Loihi, IBM TrueNorth, and SpiNNaker. 
On such platforms, energy consumption scales with synaptic event count rather than with the number of neurons or time steps, meaning that sparse spike trains translate directly into lower power draw \cite{Davies2018}.
Conventional deep neural networks executing matrix multiplications at every layer on every sample have no analogous mechanism for exploiting input sparsity at the hardware level. 
Such asymmetry motivates encoding schemes that maximize sparsity as a primary objective, treating classification accuracy as a constraint rather than the sole optimization target.

\section{Methodology} \label{sec: methodology}

\subsection{Encoder Formulations} \label{sec: encoder_methodology}
We summarize the encoder formulations in the following table.
Let $I \in [0, 1]^{H \times W}$ denote a normalized input image and $T$ the simulation time steps.
Each encoder produces a spike tensor $s \in {0,1}^{T \times D'}$ where $D'$ depends on the encoder.
    
    \begin{table}[h]
    \centering
    \caption{Summary of Evaluated Spike Encoders}
    \begin{tabular}{lll}
    \hline
    \textbf{Encoder} & \textbf{Formulation} & \textbf{$D'$} \\ \hline
    \textbf{Rate} & $s_t(p) \sim \text{Bernoulli}(I(p))$ i.i.d. for each $t \in [0,T)$ & $H \cdot W$ \\
    \textbf{Latency} & $t_{\text{spike}}(p) = \lfloor (1 - I(p))(T-1) \rfloor$, single spike per pixel & $H \cdot W$ \\
    \textbf{Delta (dual)} & \begin{tabular}[c]{@{}l@{}}jittered sequence $I_t = \text{shift}_{\Delta(t)}(I)$;\\ $s_t^\text{ON}(p) = [I_t - I_{t-1} > +\theta]$,\\ $s_t^\text{OFF}(p) = [I_t - I_{t-1} < -\theta]$\end{tabular} & $2 \cdot H \cdot W$ \\
    \textbf{DoG (dual)} & \begin{tabular}[c]{@{}l@{}}$k_\text{DoG} = G_{\sigma_c} - G_{\sigma_s}$;\\ $f^\text{ON} = [k_\text{DoG} \star I]_+$, $f^\text{OFF} = [-(k_\text{DoG} \star I)]_+$;\\ rate-code each channel\end{tabular} & $2 \cdot H \cdot W$ \\ \hline
    \end{tabular}
    \end{table}

Each of the four encoders carries its own tradeoff. \textbf{Rate coding} has the highest spike density. The same information is retransmitted at every time step, with each spike carrying little novel information. This redundancy does make the integrated signal robust to noise. \textbf{Latency coding} concentrates information into the precise timing of a single spike per neuron, approaching the theoretical minimum of one bit per spike for binary stimuli. Small timing perturbations corrupt the encoded value, and the code is unreadable until $T$ steps have elapsed. \textbf{Delta modulation} encodes change events rather than absolute values. This is efficient for signals with smooth temporal structure or structured motion, such as the microsaccade trajectory used here, where successive frames are highly correlated and the differential signal carries more information per spike than the raw signal. ON-OFF rectification splits the signed event stream into two non-negative channels. \textbf{DoG encoding} adds a spatial redundancy-reduction step prior to rate coding, removing the low-frequency correlations that dominate natural-image statistics before encoding begins. The reduction translates directly into fewer spikes per sample at matched network accuracy. As with delta, ON/OFF rectification splits the signed filter output into two channels.

Two of the four encoders, delta and DoG, double the effective input dimensionality $D'$ via dual-polarity rectification, and therefore double the parameter count of the downstream MLP's first layer relative to single-polarity encoders.
An ON/OFF-split encoder hands the network a wider, more informative representation at the cost of more trainable parameters for the same network architecture and hidden layer width.

\begin{figure}[h]
    \centering
    \includegraphics[width = \textwidth]{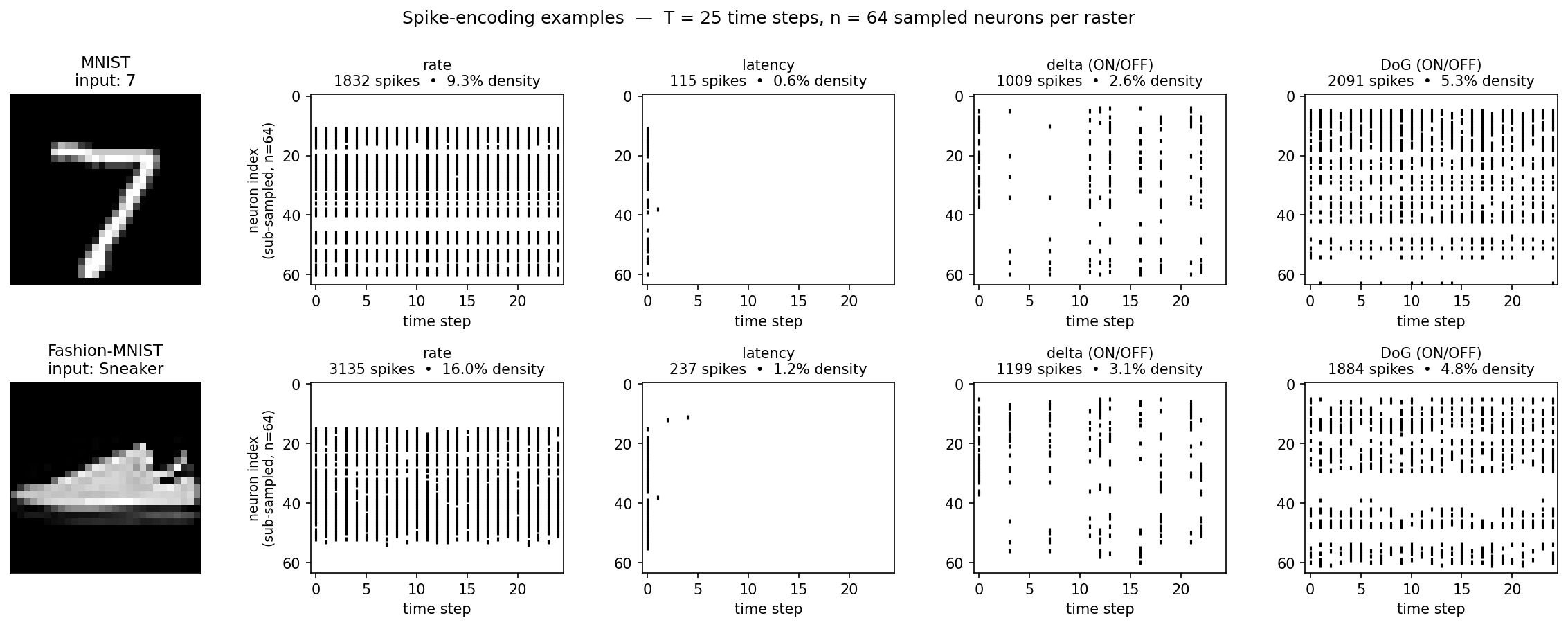}
    \caption{Example spike trains generated from the four encoders on MNIST digit and Fashion-MNIST datasets}
    \label{fig:encoding-diagram}
\end{figure}

Figure \ref{fig:encoding-diagram} illustrates how the same input, one MNIST digit and one Fashion-MNIST, is converted into spike trains by each of the four encoders. 
Each panel is a spike raster of $T = 25$ timesteps (horizontal axis) over a 64-neuron subset(vertical axis), generated using snntorch.spikeplot.raster.

\subsection{Network Architecture} \label{sec: network_methodology}

Our core SNN is a three-layer feedforward network (SpikingMLP) with
fully-connected layers mapping \(D\) input dimensions through hidden
layers of width 256 and 128 before projecting to \(C\) output classes.
The first layer applies a SparseLinear transformation from \(D\) to 256
followed by a LIF nonlinearity, the second applies SparseLinear from 256
to 128 followed by LIF, and the third applies a standard Linear
projection from 128 to \(C\) followed by LIF. The SparseLinear layer
applies a fixed random binary mask to the weight matrix:

\begin{equation}
\mathbf{y} = (\mathbf{W} \odot \mathbf{M})\mathbf{x}, \quad M_{ij} \sim \text{Bernoulli}(p_{\text{conn}})
\end{equation}

where \(p_{\text{conn}} = 1.0\) controls the connectivity fraction
(fully connected by default), but sparsity can be induced to reduce
computation.

For comparison, we also train a traditional neural network (NN), a
convolutional neural network (CNN), and a zero-shot converted SNN
(ConvertedSNN), in which the weights of a trained NN are directly
transplanted into an LIF architecture by replacing ReLU activations with
LIF neurons and removing dropout.

Three readout modes are supported. The spike count readout computes
\(\hat{y} = \sum_{t=1}^T s_{\text{out}}(t)\) and is appropriate for
dense (rate) codes. The membrane sum readout computes
\(\hat{y} = \sum_{t=1}^T u_{\text{out}}(t)\) and is appropriate for
sparse (latency and delta) codes. The final membrane readout computes
\(\hat{y} = u_{\text{out}}(T)\) and is appropriate for late-spike codes.

\subsection{Weight Initialization and Connectivity}

Weight initialization has outsized importance in SNNs relative to
conventional networks because the spiking nonlinearity is not
scale-invariant. A weight matrix initialized too large will drive all
neurons to fire on every time step (saturating the code), while one
initialized too small will produce silence. We initialize weights using
Kaiming uniform initialization scaled by the inverse square root of the
fan-in, which maintains stable firing rates near 0.1 spikes per neuron
per time step at initialization. The sparse connectivity mask is fixed
after initialization and is not updated during training, so the
effective fan-in for each neuron is
\(p_{\text{conn}} \times d_{\text{in}}\) .

\subsection{Surrogate Gradient and Rouser Adaptive Threshold Learning}
\label{sec:surrogate_rouser_methodology}

Direct backpropagation through an SNN is difficult because the spike function is a hard threshold:
\begin{equation}
s(t)=\Theta(u(t)-\vartheta),
\end{equation}
where \(u(t)\) is the membrane potential and \(\vartheta\) is the firing threshold. The forward rule is easy to read, a neuron fires once its membrane potential crosses the threshold. The trouble appears in the backward pass. The derivative of \(\Theta(\cdot)\) is zero almost everywhere and undefined at the threshold, so direct gradient descent receives little usable signal through the spike output.

Surrogate-gradient learning handles this by keeping the hard binary spike in the forward pass and replacing its derivative with a smooth approximation in the backward pass. For a fixed-threshold SNN, this gives
\begin{equation}
\frac{\partial s}{\partial u}
\approx
\phi(u-\vartheta).
\end{equation}
In our baseline SNN, we use the fast-sigmoid surrogate gradient from snntorch:
\begin{equation}
\tilde{\sigma}'(u-\vartheta)
=
\frac{k}{(1+k|u-\vartheta|)^2},
\qquad k=25.
\end{equation}
The spike computation remains discrete, while the optimizer receives a local slope near the firing boundary \cite{Neftci2019,Zenke2018}.

Rouser extends the same surrogate-gradient framework by making the threshold learnable. A standard surrogate-gradient SNN learns synaptic weights with a fixed threshold. Rouser learns both weights and thresholds, allowing each neuron to tune its firing sensitivity during training \cite{Takaghaj2025}. Its forward spike rule remains
\begin{equation}
s=\Theta(u-\vartheta),
\end{equation}
and the backward pass assigns surrogate gradients to both \(u\) and \(\vartheta\):
\begin{equation}
\frac{\partial s}{\partial u}
=
\frac{s_0}{\tau_r}
\exp\!\left(
-\frac{|u-\vartheta|}{\tau_r}
\right),
\qquad
\frac{\partial s}{\partial \vartheta}
=
-\frac{s_0}{\tau_r}
\exp\!\left(
-\frac{|u-\vartheta|}{\tau_r}
\right).
\end{equation}
The two gradients have the same shape and opposite signs. Increasing membrane potential makes a spike more likely; increasing the threshold makes a spike harder to trigger.

The threshold update is then
\begin{equation}
\vartheta
\leftarrow
\vartheta
-
\eta_{\vartheta}
\frac{\partial \mathcal{L}}{\partial \vartheta}.
\end{equation}
This gives the model a second adjustment path. The weights determine how much membrane potential is accumulated, and the threshold determines when that accumulated potential becomes a spike. In practice, this can help rouse inactive neurons whose membrane potentials remain too far below threshold. It also makes \(\eta_{\vartheta}\) important. A small value behaves close to a fixed-threshold SNN, while a large value can move thresholds too aggressively and make spike dynamics less stable. We evaluate this threshold learning-rate effect in Section~\ref{sec:rouser_results}.

\subsection{Adam Optimization}

Adam stands for Adaptive Moment Estimation, is a  first-order stochastic optimization method, commonly used for training neural networks. Unlike the standard gradient descent method, which only uses the current gradient to update the parameters, Adam maintains two exponential moving averages simultaneously. The first average $m_t$ is used to estimate the average value of the gradient direction, and the second average $v_t$ is used to estimate the average squared value of the gradient magnitude\cite{kingma2014adam}.

\begin{equation}
g_t \leftarrow \nabla_{\theta} f_t(\theta_{t-1})
\end{equation}
\begin{equation}
m_t \leftarrow \beta_1 m_{t-1} + (1-\beta_1) g_t, \quad  v_t \leftarrow \beta_2 v_{t-1} + (1-\beta_2) g_t^2
\end{equation}

The $g_t$ represents the gradient at step $t$. $\beta_1$ and $\beta_2$ represent exponential decay rates for the moment estimates. In our SNN experiments, Adam was used as the main first-order benchmark optimizer for training the alternative gradient spiking neural network. Since SNN contains non-differentiable spike activation functions, we used alternative gradients during the backpropagation process, enabling the model to be trained using standard gradient-based optimizers. Adam is highly applicable as it is specifically designed for stochastic optimization and uses adaptive estimates of the first and second moments of the gradients, making it relatively stable even in the presence of noisy gradients.

\subsection{Limitations of Adam}

As Adam is a first-order optimizer, it does not explicitly model the curvature of the loss landscape and only uses gradient information. Although the surrogate gradient makes backpropagation possible, this gradient is only an approximation of the true spike derivative. Adam can smooth this noisy gradient by using the moving average method, but it cannot correct the mismatch between the hard spike function used in the forward propagation and the surrogate derivative used in the backpropagation.

\subsection{K-FAC Optimization}

Standard first-order optimizers (SGD, Adam) use only gradient
information. Kronecker-Factored Approximate Curvature (K-FAC) \cite{Martens2015}
incorporates second-order curvature information through a tractable
approximation to the Fisher Information Matrix.

Before we introduce this optimization method, it is worth noting that K-FAC is an optimization method designed to make neural network training more efficient by approximating the natural gradient. In standard gradient descent, the formula of standard gradient can be described as follows:

\begin{equation}
\theta_{t+1} = \theta_t - \eta \nabla_{\theta} L
\end{equation}

where L is the loss function, $\theta$ represents the model parameters, and $\eta$ is the learning rate. However, the standard decent gradient does not account for the geometry of the model’s output distribution, which can make optimization slow or poorly conditioned. For better performance, Natural gradient methods improve this by using F which is the Fisher Information Matrix. This approach captures curvature information about the model. The natural-gradient update is:

\begin{equation}
\theta_{t+1} = \theta_t - \eta F^{-1}\nabla_{\theta}L
\end{equation}

This update rescales the gradient using curvature information, allowing the optimizer to take more meaningful steps in parameter space. It is computationally expensive to store and invert Fisher Information Matrix because of its extremely larger size for neural network. K-FAC is able to solve this issue by approximating the Fisher matrix in a structured and efficiently invertible form.

For a linear layer with weight matrix
\(\mathbf{W} \in \mathbb{R}^{d_{\text{out}} \times d_{\text{in}}}\), the
exact Fisher block is:

\begin{equation}
F_\ell = \mathbb{E}\!\left[\mathbf{g} \mathbf{g}^\top \otimes \mathbf{a} \mathbf{a}^\top\right]
\end{equation}

where \(\mathbf{a} \in \mathbb{R}^{d_{\text{in}}}\) is the layer's input
activation and \(\mathbf{g} \in \mathbb{R}^{d_{\text{out}}}\) is the
gradient of the loss with respect to the layer output.
K-FAC approximates this as
\(F_\ell \approx \mathbf{G} \otimes \mathbf{A}\), where
\(\mathbf{A} = \mathbb{E}[\mathbf{a}\mathbf{a}^\top]\) and
\(\mathbf{G} = \mathbb{E}[\mathbf{g}\mathbf{g}^\top]\) are computed
separately. This factorization makes inversion tractable via the
Kronecker identity
\((\mathbf{G} \otimes \mathbf{A})^{-1} = \mathbf{G}^{-1} \otimes \mathbf{A}^{-1}\) \cite{Martens2015}.

Within every pre-designed size batches, \(\mathbf{A}\) and
\(\mathbf{G}\) are updated as exponential moving averages with factor
decay \(\rho = 0.95\) \cite{Martens2015}:

\begin{equation}
\mathbf{A} \leftarrow \rho \mathbf{A} + (1-\rho) \frac{1}{B} \mathbf{a}_{\text{aug}}^\top \mathbf{a}_{\text{aug}}, \quad \mathbf{G} \leftarrow \rho \mathbf{G} + (1-\rho) \frac{1}{B} \mathbf{g}^\top \mathbf{g}
\end{equation}
Here, \(\mathbf{a}_{\text{aug}}\) is the activation matrix augmented with a column of ones to include the bias parameters. To enhance numerical stability, each factor is inverted by using the Tikhonov regularization method based on the Cholesky decomposition with initial damping \(\lambda = 10^{-3}\). The damping increases exponentially (by \(10\times\)) up to six times if
Cholesky fails, with a fallback to the Moore-Penrose pseudo-inverse. The
natural gradient update for the weight matrix is:

\begin{equation}
\Delta \mathbf{W} = -\eta \cdot \mathbf{G}^{-1} \nabla_{\mathbf{W}} \mathcal{L} \cdot \mathbf{A}^{-1}
\end{equation}

In some situation, if A or G is noisy, nearly singular, or poorly estimated, then their inverses can make the update extremely large. Therefore, we applied a clipping function to our weight matrix: 

\begin{equation}
\Delta \tilde{W} \leftarrow \Delta \tilde{W}\cdot \min\left(1,\frac{r}{\left\lVert \Delta \tilde{W} \right\rVert}\right)
\end{equation}

This will avoid negative impacts including unstable loss, poor accuracy and Null values due to the large difference after weight updating.

K-FAC proved to be useful as it enabled investigations into how curvature-aware optimization affects the training behavior of neural networks and spiking neural networks. It conducted a meaningful comparison with Adam by demonstrating how second-order information influences convergence speed, stability, and final accuracy. For spiking neural networks, K-FAC is more powerful because it creates a differentiable approximation of the spiking model by replacing gradient training, thus making curvature-aware optimization possible.

\subsection{K-FAC Across Time Steps}

One known limitation of applying K-FAC to recurrent computations is that its Fisher decomposition assumes that the input activations and output gradients are independent of each other. However, when the same weight matrix is applied in multiple time steps, and the gradients in subsequent time steps are dependent on the activations from earlier time steps through time backpropagation, this assumption is violated. 
In our implementation, we only apply K-FAC to the spatial weight matrix, accumulating the activation and gradient statistics along the time dimension, as if the time steps were independent samples. This approximation is equivalent to treating time steps as independent samples in a batch dimension, which underestimates the true Fisher curvature. However, in shallow SNNs \cite{Martens2015, Guo2023}, this is computationally feasible and performs well in practical applications. A completely correct approach requires performing Fisher decomposition on time-dependent structures, but for LIF networks, this remains an unsolved research problem.

\section{Experimental Setup}

\subsection{Datasets}
All experiments use an 80/20 train/test split with no separate
validation set and a fixed random seed of 0 for reproducibility. MNIST is loaded from CSV, normalized to $[0,1]$, and flattened into $784$-dimensional vectors; the 60,000-sample training set and 10,000-sample test set are re-split at the 80/20 ratio. 
Fashion-MNIST is loaded via torchvision with the official 60,000/10,000 split similarly re-partitioned. 

\subsection{Encoder Hyperparameters}
Two hyperparameters must be set empirically before comparisons across encoders are meaningful. The first is the simulation length $T$ per encoder, and the second is the DoG center Gaussian $\sigma_c$.

Each encoder uses the time axis differently, as described in Section \ref{sec: encoder_methodology}, which leads to a different optimal $T$ per encoder. These values are shown in Table~\ref{tab: T}. The optimal $\sigma_c$ also varies between datasets, as shown in Table~\ref{tab: sigma}. Fashion-MNIST silhouettes discriminate via fine high-frequency edges such as collars and hems, and prefer sub-pixel center kernels. MNIST discriminates via stroke geometry, requiring $\sigma_c$ to exceed the stroke half-width so the surround Gaussian extends beyond the stroke and the DoG produces a strong response. For canonical comparisons we use one $\sigma_c$ per dataset, taking the value optimal at the parameter-Pareto sweet spot $h = 32$ (Fashion-MNIST: 0.5; MNIST: 1.5).

We compare four spike encoders, rate, latency, delta (dual polarity), and DoG (dual polarity), on two image-classification datasets (Fashion-MNIST, MNIST) using the fixed three-layer feedforward SpikingMLP (see Section \ref{sec: network_methodology}) with a vanilla Adam optimizer.
The baseline spiking MLP uses hidden layer widths $(D' \to h \to h/2 \to 10)$, with hidden width $h$ as the primary swept axis ($h \in \{4, 8, 16, 32, 64, 128, 256\}$) to evaluate performance per parameter.

Tables~\ref{tab:fmnist_accuracy} and \ref{tab:mnist_accuracy} summarize the combination of simulation duration $T$ and center scale $\sigma_c$ that yielded peak absolute accuracy for each hidden layer width $h$ across both Fashion-MNIST and MNIST.

\begin{table}[H]
\centering
\caption{Optimal Simulation Time ($T$) by Encoder}
\label{tab: T}
\begin{tabular}{lll}
\hline
\textbf{Encoder} & \textbf{Optimal $T$} & \textbf{Reason} \\ \hline
\textbf{Rate} & $T = 50$ (still climbing) & Linear spike-count scaling \\
\textbf{DoG} & $T = 25$ (plateau) & Linear scaling, accuracy plateau at $T=25$ \\
\textbf{Latency} & $T = 10$ & One spike per pixel; longer $T$ adds empty time \\
\textbf{Delta} & $T = 10$ & Bounded saccade-trajectory event count \\ \hline
\end{tabular}
\end{table}

\begin{table}[H]
\centering
\caption{Optimal $\sigma_c$ per (dataset, $h$)}
\label{tab: sigma}
\label{tab:dataset_h_comparison}
\begin{tabular}{lllllll}
\hline
\textbf{Dataset} & \textbf{$h=8$} & \textbf{$h=16$} & \textbf{$h=32$} & \textbf{$h=64$} & \textbf{$h=128$} & \textbf{$h=256$} \\ \hline
Fashion-MNIST    & —              & —               & 0.5             & 1.0             & 0.5              & 0.5              \\
MNIST            & 0.5            & 1.0             & 1.5             & 1.5             & 1.5              & 2.0              \\ \hline
\end{tabular}
\end{table}

\begin{table}[H]
\centering
\caption{Fashion-MNIST Accuracy Comparison (Rate $T=50$; Latency $T=10$; Delta $T=10$; DoG $\sigma_c=0.5, T=25$)}
\label{tab:fmnist_accuracy}
\begin{tabular}{rlllll}
\hline
\textbf{$h$} & \textbf{Rate} & \textbf{Latency} & \textbf{Delta} & \textbf{DoG} & \textbf{DoG $\Delta$ vs. Rate} \\ \hline
8   & 75.2 & 73.0 & 74.2 & \textbf{82.4} & +7.2 \\
16  & 82.0 & 79.4 & 80.3 & \textbf{84.7} & +2.7 \\
32  & 85.0 & 80.7 & 83.9 & \textbf{86.6} & +1.6 \\
64  & 84.3 & 82.3 & 84.5 & \textbf{86.8} & +2.4 \\
128 & 85.3 & 80.8 & 83.3 & \textbf{87.4} & +2.1 \\
256 & 84.2 & 82.7 & 83.9 & \textbf{87.9} & +3.6 \\ \hline
\end{tabular}
\end{table}

\begin{table}[H]
\centering
\caption{MNIST Accuracy Comparison (All encoders $T=25$; DoG $\sigma_c=1.5$)}
\label{tab:mnist_accuracy}
\begin{tabular}{rlllll}
\hline
\textbf{$h$} & \textbf{Rate} & \textbf{Latency} & \textbf{Delta} & \textbf{DoG} & \textbf{DoG $\Delta$ vs. Rate} \\ \hline
8   & \textbf{78.3} & 46.8 & 37.4 & 62.5 & -15.8 \\
16  & \textbf{84.8} & 65.8 & 75.7 & 83.6 & -1.2  \\
32  & 91.8          & 77.7 & 87.7 & \textbf{93.6} & +1.9  \\
64  & 92.9          & 84.3 & 86.1 & \textbf{94.1} & +1.3  \\
128 & 94.4          & 85.0 & 87.7 & \textbf{94.8} & +0.4  \\
256 & 94.9          & 87.8 & 89.7 & \textbf{95.1} & +0.2  \\ \hline
\end{tabular}
\end{table}

\subsection{Network Hyperparameters}
All model families share a common set of base hyperparameters. Training
is performed for 10 epochs with a mini-batch size of 64. The base
learning rate is \(\eta_w = 10^{-3}\) using the Adam optimizer with
default PyTorch settings (\(\beta_1 = 0.9\), \(\beta_2 = 0.999\),
\(\epsilon = 10^{-8}\)). For Rouser, thresholds are optimized through a separate parameter group. We sweep
\(\eta_\vartheta \in \{0,10^{-3},5\times10^{-3},10^{-2}\}\), where
\(\eta_\vartheta=0\) corresponds to the fixed-threshold baseline. The best single-run Rouser setting is
\(\eta_\vartheta=5\times10^{-3}\). All SNN simulations run for \(T = 25\) time steps with LIF
membrane decay \(\beta = 0.9\) and fixed initial spike threshold
\(\vartheta = 1\).

Each encoder also requires its own parameters. Latency coding uses
\(\tau = 1.0\) and a minimum-activation threshold of
\(\vartheta_{\text{lat}} = 0.01\). Neurons with
\(p_i \leq \vartheta_{\text{lat}}\) do not fire, and spike times
exceeding \(T - 1\) are suppressed. Delta modulation uses a change
threshold of \(\delta = 0.1\) and a saccade amplitude of \(A = 2\)
pixels. DoG encoding uses \(\sigma_c = 0.5\) and \(\sigma_s = 1.0\).
K-FAC uses an initial damping of \(\lambda = 10^{-3}\), a factor decay
of \(\rho = 0.95\), and curvature updates every 10 mini-batches.

\subsection{Hardware and Software}
All experiments were run on a single NVIDIA GeForce RTX 3060 GPU under
Python 3.12. The implementation uses PyTorch and snntorch with no
version constraints beyond those required by Python 3.12 compatibility.
No multi-GPU or distributed training is employed.

\section{Results} \label{sec: results}

\subsection{Parameter Efficiency Experiment: Spike Encoding as a Parameter-Efficiency Lever}

We compare four spike encoders on two image-classification datasets (Fashion-MNIST, MNIST) under a fixed three-layer feedforward SpikingMLP with vanilla Adam optimizer.
The architecture, training procedure, and hardware are identical across encoders; encoder choice is the only experimental variable. 
We use the vanilla Adam optimizer at $\text{lr} = 10^{-3}$, batch size 64, $80\%$ train / $20\%$ test split, 5 epochs. 
Two seeds per configuration used in reported error bars. 
Where the rest of the paper studies network-side optimization, this section isolates the contribution of the encoder itself.

\begin{figure}[H]
    \centering
    \includegraphics[width = \textwidth]{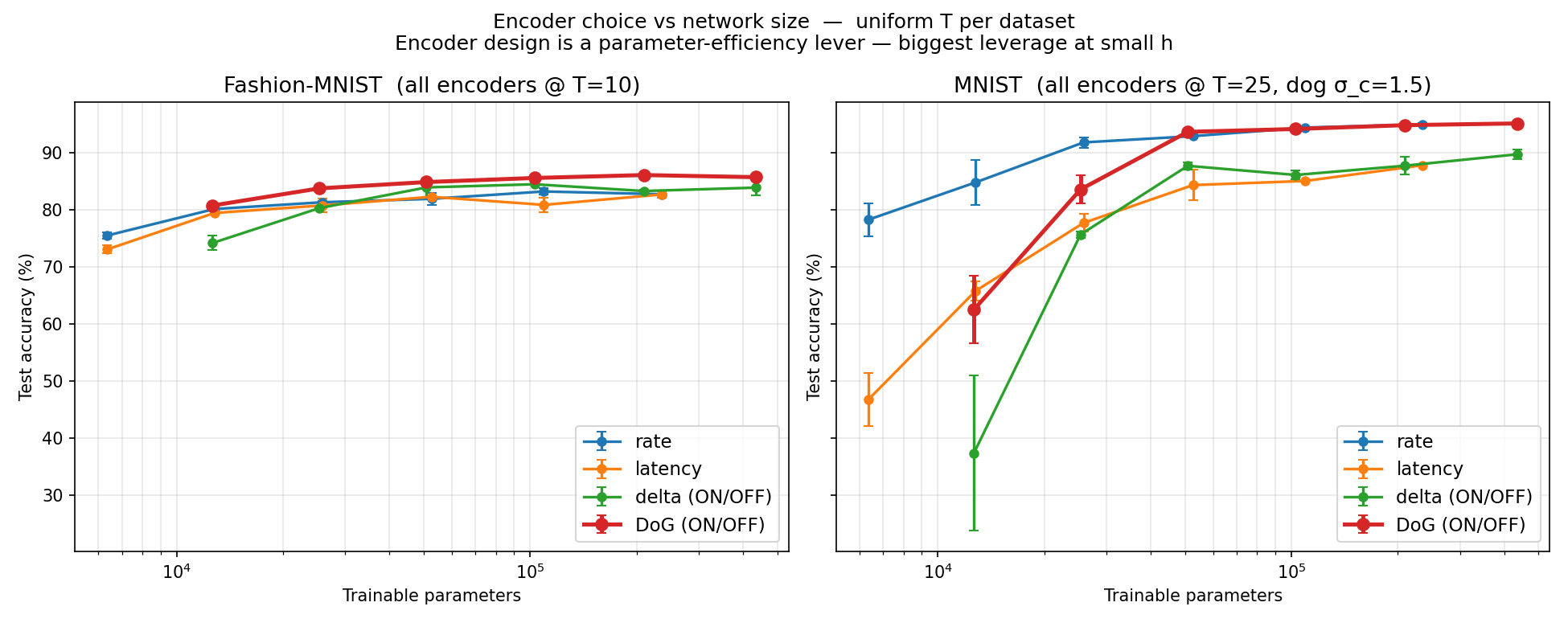}
    \caption{Test accuracy vs trainable parameters for four spike encoders on a simple SNN MLP}
    \label{fig:encoding-result}
\end{figure}

Figure \ref{fig:encoding-result} shows test accuracy against trainable parameter count on Fashion-MNIST and MNIST for all four encoders, with $h$ swept across range $h \in {4, 8, 16, 32, 64, 128, 256})$.
This generates for each encoder a curve from 6k parameters to 436k parameters, with dual-polarity encoders at any given $h$ requiring twice the first hidden layer weights than single-polarity encoders. 
For control purposes, Fashion-MNIST uses uniform $T = 10$ across all four encoders and MNIST uses uniform $T = 10$; holding $T$ constant compares encoders at matched simulation-time budget. 

Figure \ref{fig:encoding-result} shows that DoG encoders hit a sweet spot at $h = 32$. Beyond that point, returns to capacity diminish faster than returns to encoder choice. The interpretation is that a sufficiently-wide MLP trained on enough samples can learn a DoG-like spatial filter from data, which makes the explicit DoG kernel redundant for raw accuracy. To reach DoG at $h = 32$ accuracy with rate coding requires $h \ge 64$ on MNIST and is unachievable on Fashion-MNIST at any tested $h$. On Fashion-MNIST relative to delta, DoG shows a 4$\times$ parameter saving at matched accuracy (DoG at $h = 16$ with 25k params $\approx$ delta at $h = 64$ with 103k params) and an 8.6$\times$ parameter saving at $+2.7$ pp higher accuracy (DoG at $h = 32$ with 51k params versus delta at $h = 256$ with 436k params). The accuracy gap between encoders is large at small parameter counts and narrows as $h$ grows, and the encoder advantage is recoverable by scaling the model, but at significant parameter cost.

We conclude that encoder choice is not an absolute-accuracy level, but a parameter-efficiency lever.
Its leverage is largest when facing the energy consumption and network size constraint, which is exactly where biological neural systems and low-power neuromorphic hardware operate.

\subsection{Network Optimizer Experiment}

\subsubsection{Initial Neural Network Model}
We evaluate four model families on MNIST and Fashion-MNIST. Key metrics include test
accuracy, test loss, training accuracy, training loss, parameter count, model
size, and peak memory usage.

Before applying optimizer and using the surrogate-gradient, we want to see the performance of our initial spike neural network:

\begin{figure}[H]
    \centering
    \includegraphics[width = 0.8\textwidth]{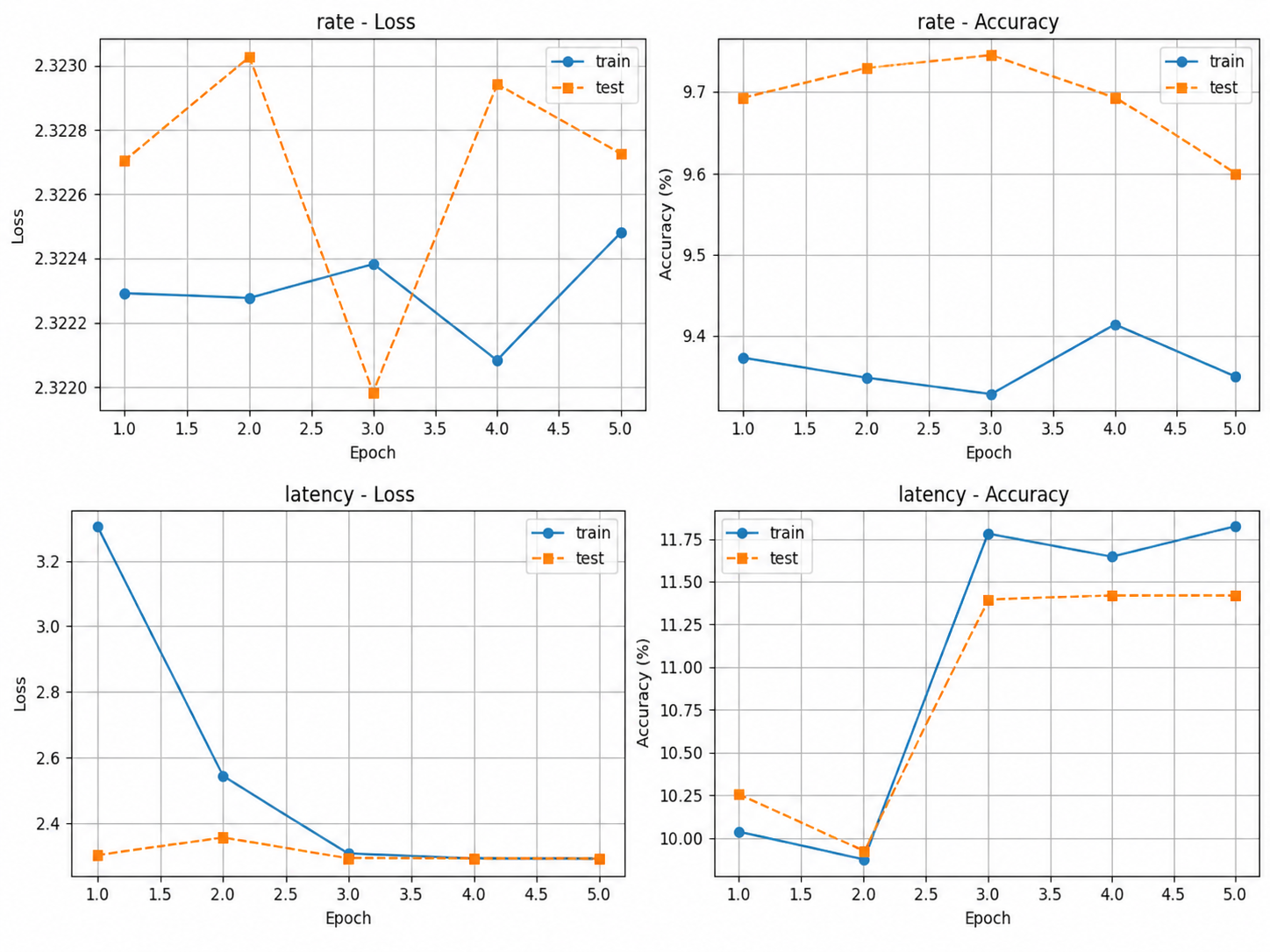}
    \caption{Training accuracy/loss vs. Testing accuracy/loss for rate and latency encoding }
    \label{fig:rate_latency_ini}
\end{figure}

\begin{figure}[H]
    \centering
    \includegraphics[width = 0.8\textwidth]{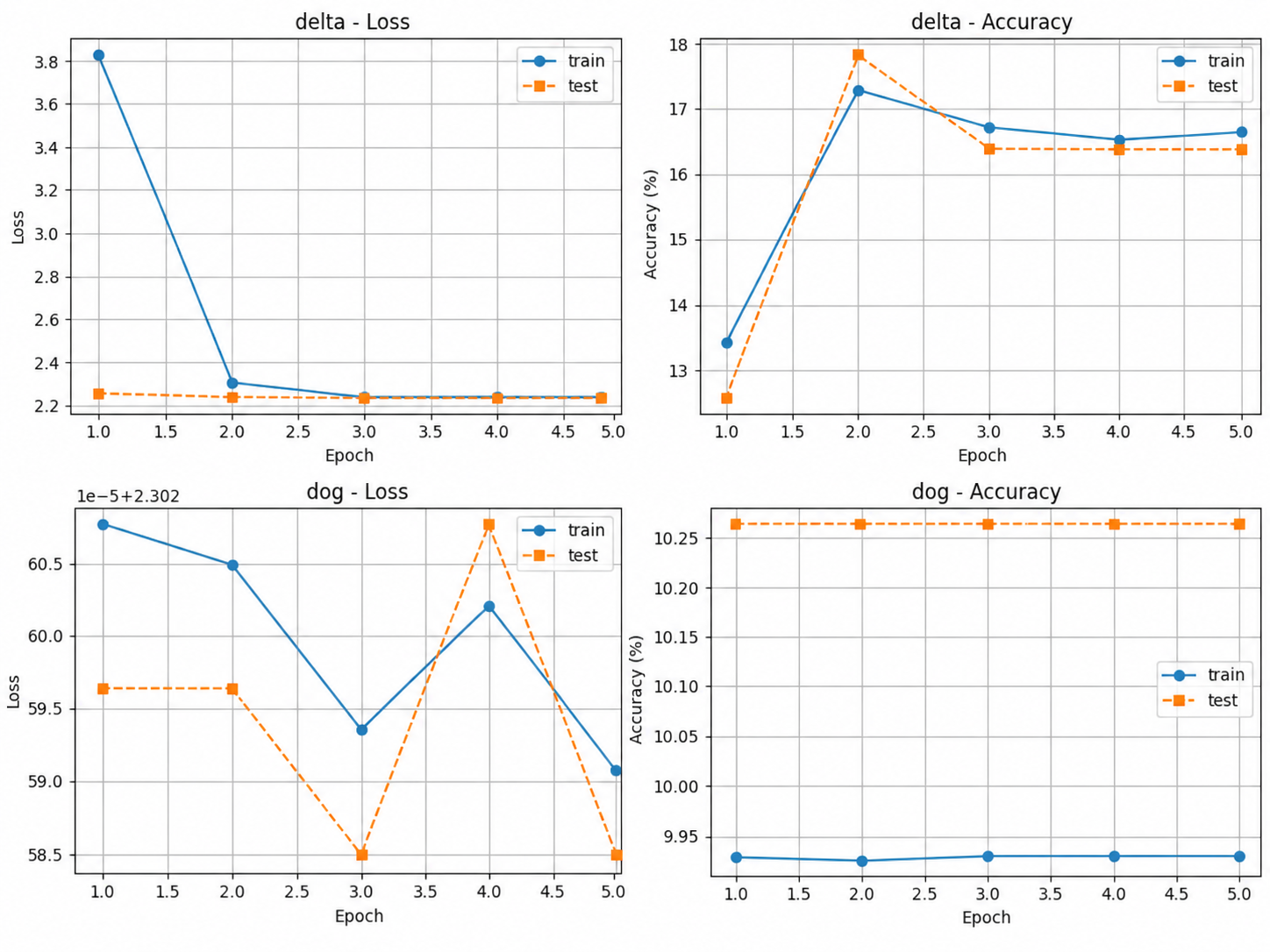}
    \caption{Training accuracy/loss vs. Testing accuracy/loss for delta and dog encoding }
    \label{fig:delta_dog_ini}
\end{figure}

Figure \ref{fig:rate_latency_ini} and Figure \ref{fig:delta_dog_ini} show the baseline behavior of the SNN before applying the optimizer improvements such as K-FAC and before using the surrogate-gradient training method effectively. Overall, these results indicate that the model has not learned meaningful features for distinguishing different categories. Most of the curves in these figures stay close to 10\% to 18\%, which suggests that the network output is only slightly better than random and that the training signal is weak or unstable. The DoG encoding result is the weakest. The accuracy stays almost constant around 10\%, and the loss remains close to the random-guessing region.

\subsubsection{Rouser Adaptive Threshold Results}
\label{sec:rouser_results}

We also evaluate Rouser on Fashion-MNIST using the same three-layer SNN backbone. This experiment asks a narrow question: does a learnable threshold improve a simple surrogate-gradient SNN compared with a fixed-threshold version? The input images are converted into spike trains using internal Bernoulli rate coding. For each normalized pixel value \(x \in [0,1]\), the input spike at time \(t\) is sampled as
\begin{equation}
x_t \sim \mathrm{Bernoulli}(x).
\end{equation}
The model runs for \(T=25\) time steps, and the predicted class is selected by accumulated output spike count:
\begin{equation}
\hat{y}
=
\arg\max_c
\sum_{t=1}^{T} s_c(t).
\end{equation}

\begin{table}[H]
\centering
\caption{Threshold learning-rate sweep on Fashion-MNIST.}
\label{tab:rouser_lr_sweep}
\begin{tabular}{llll}
\hline
\textbf{Threshold LR \(\eta_\vartheta\)} & \textbf{Best test accuracy} & \textbf{Best epoch} & \textbf{Observation} \\
\hline
0 & 87.37\% & 7 & Fixed-threshold baseline \\
\(10^{-3}\) & 87.60\% & 9 & Small improvement \\
\(5\times10^{-3}\) & \textbf{88.13\%} & 9 & Best accuracy and stable thresholds \\
\(10^{-2}\) & 87.78\% & 10 & Good accuracy, less stable thresholds \\
\hline
\end{tabular}
\end{table}

The sweep shows that adaptive threshold learning improves over the fixed-threshold baseline in the best single-run results. The best setting is
\(\eta_\vartheta=5\times10^{-3}\), which reaches 88.13\% test accuracy. The larger value, \(10^{-2}\), also performs well, although the learned first-layer thresholds become less stable in our run. This matches the role of \(\eta_\vartheta\) in the update rule because it directly scales the new threshold-gradient path.

\begin{figure}[H]
    \centering
    \includegraphics[width=0.85\textwidth]{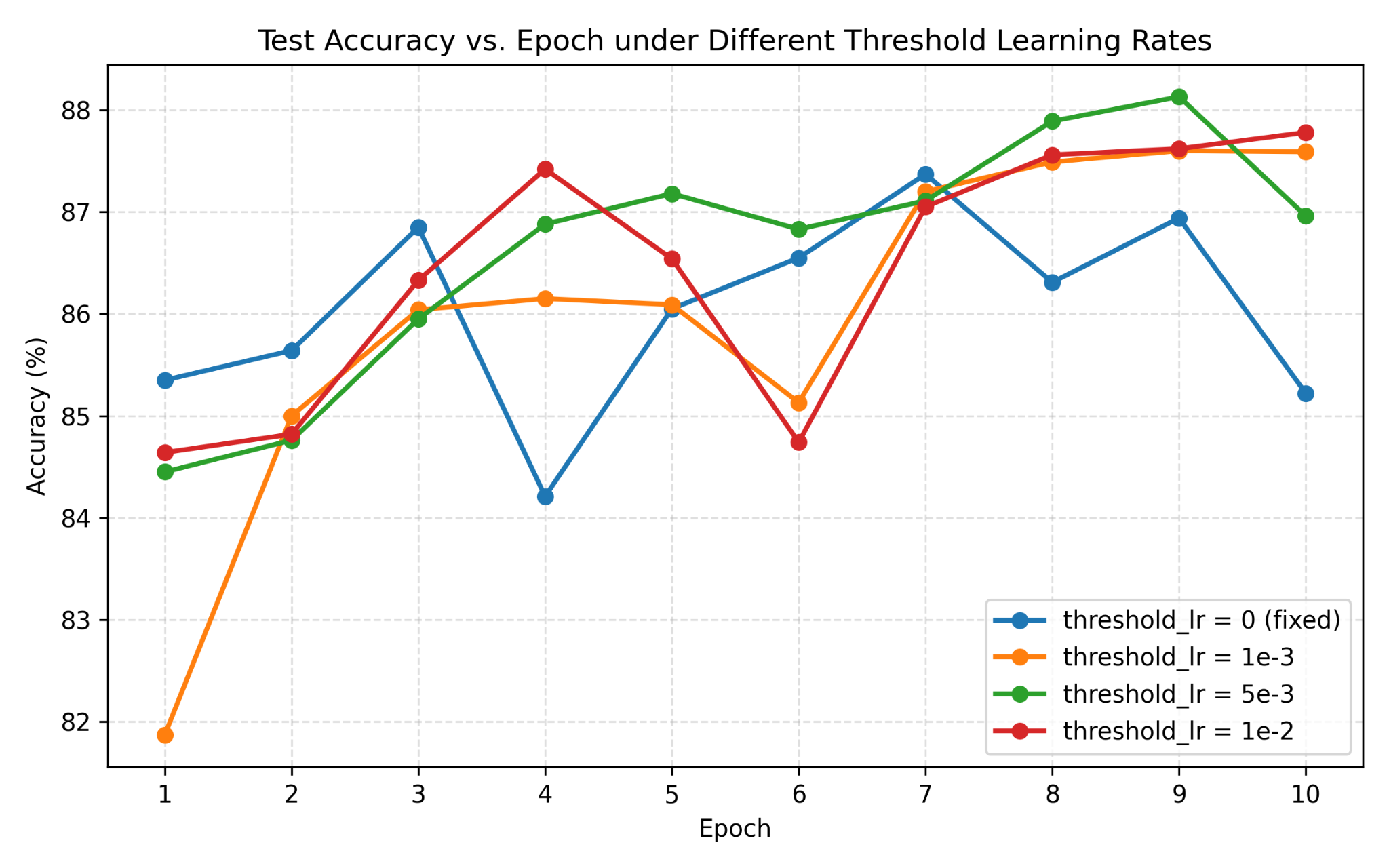}
    \caption{Test accuracy over epochs under different threshold learning rates.}
    \label{fig:rouser_lr_curves}
\end{figure}

Because Bernoulli rate coding introduces randomness, we repeat the key comparison across three random seeds. We compare the fixed-threshold baseline with the best Rouser setting from the sweep:
\begin{equation}
\eta_\vartheta \in \{0,5\times10^{-3}\}.
\end{equation}

\begin{table}[H]
\centering
\caption{Multi-seed comparison between fixed threshold and Rouser.}
\label{tab:rouser_seed_repeat}
\begin{tabular}{lllll}
\hline
\textbf{Setting} & \textbf{Seeds} & \textbf{Mean best acc.} & \textbf{Std} & \textbf{Mean best loss} \\
\hline
Fixed threshold, \(\eta_\vartheta=0\) & 3 & 87.44\% & \(\pm 0.43\%\) & 0.3686 \\
Rouser, \(\eta_\vartheta=5\times10^{-3}\) & 3 & \textbf{87.75\%} & \(\pm 0.32\%\) & \textbf{0.3556} \\
\hline
\end{tabular}
\end{table}

The multi-seed result is more conservative than the best single run, yet the direction stays positive. Rouser improves mean best accuracy from 87.44\% to 87.75\%, and the mean best loss drops from 0.3686 to 0.3556. The gain is modest, which is expected because Fashion-MNIST is a static image dataset rather than a native event-based dataset. Still, the result supports the main mechanism: adaptive thresholds give the SNN a small but measurable optimization benefit.

\subsubsection{Adam $\&$ K-FAC}

Next, we will try applying surrogate gradient method into our SNN model. At this time, we will use Adam optimization to improve the performance of our model. The results are shown below:

\begin{figure}[H]
    \centering
    \includegraphics[width = 0.8\textwidth]{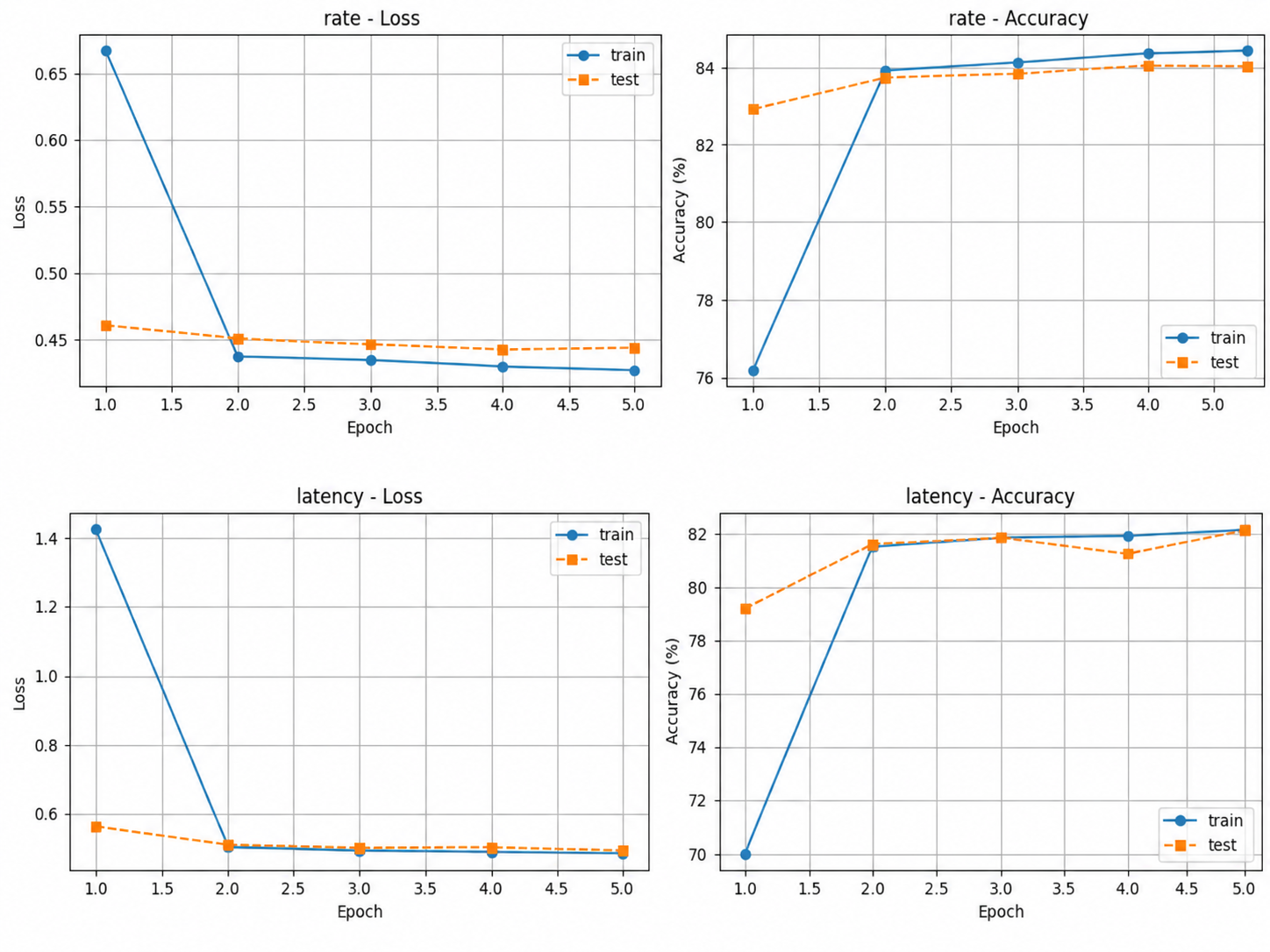}
    \caption{Training accuracy/loss vs. Testing accuracy/loss for rate and latency encoding (Adam)}
\end{figure}

\begin{figure}[H]
    \centering
    \includegraphics[width = 0.8\textwidth]{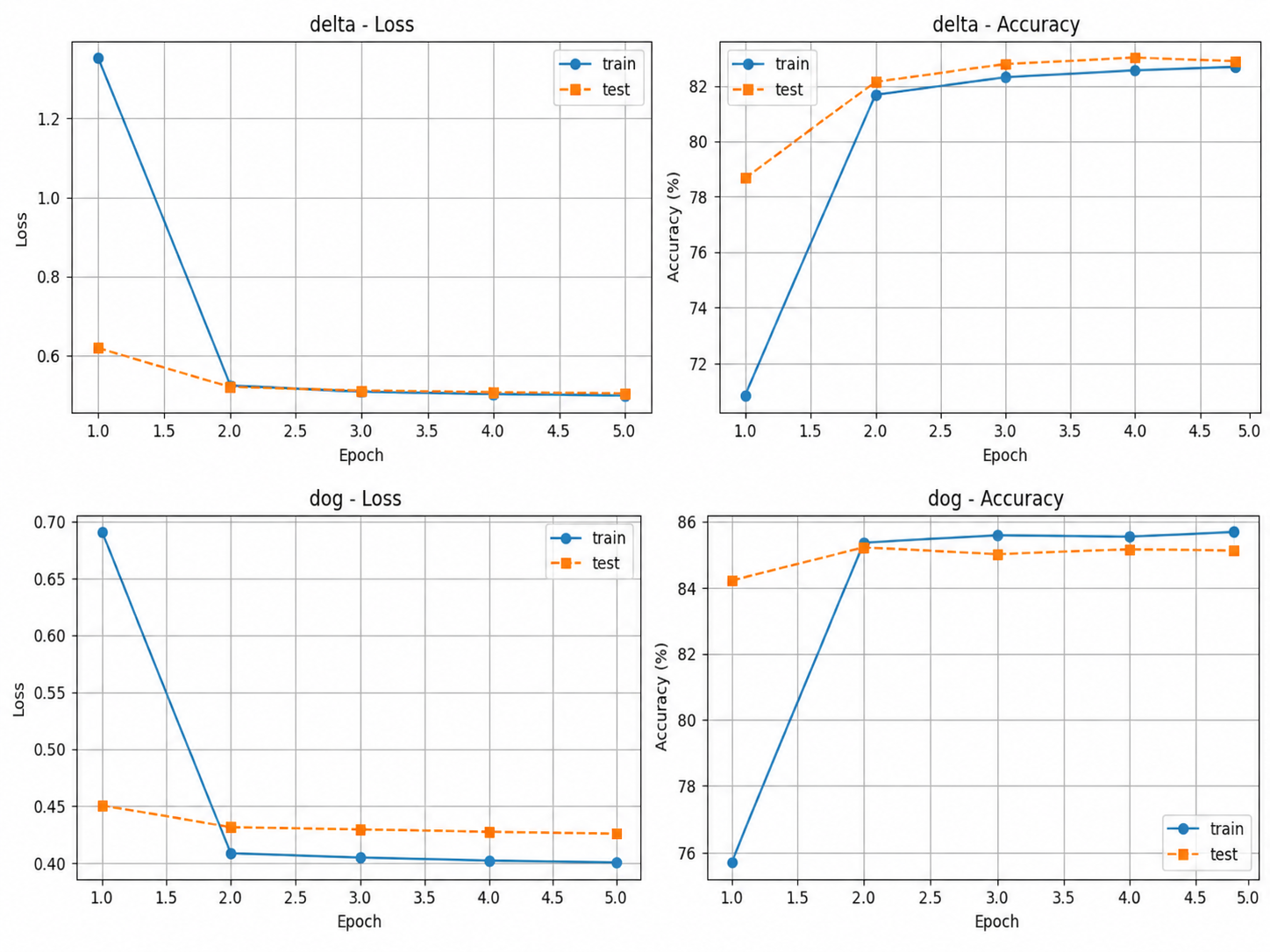}
    \caption{Training accuracy/loss vs. Testing accuracy/loss for delta and dog encoding (Adam)}
\end{figure}

These curves show a clear improvement after applying surrogate-gradient training and Adam optimization. Compared to the previous result, the network now reaches roughly 80–85\% accuracy for several encoding methods. Once the hard spike function is trained using surrogate gradients and Adam is used to stabilize the noisy updates, the SNN begins to learn meaningful class-discriminative representations. Next we will change the optimization method from Adam to K-FAC:

\begin{figure}[H]
    \centering
    \includegraphics[width = 0.8\textwidth]{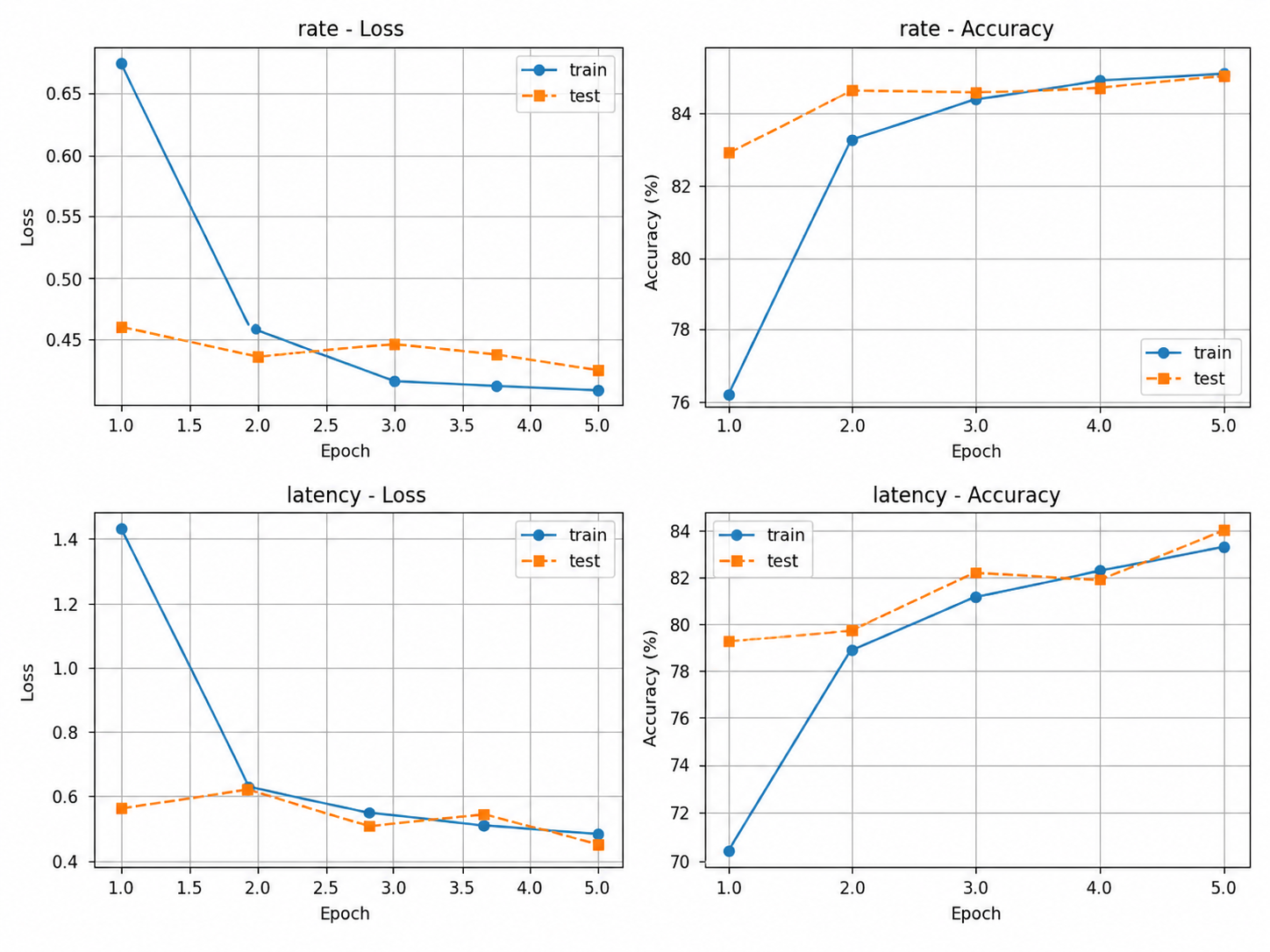}
    \caption{Training accuracy/loss vs. Testing accuracy/loss for rate and latency encoding (K-FAC)}
\end{figure}

\begin{figure}[H]
    \centering
    \includegraphics[width = 0.8\textwidth]{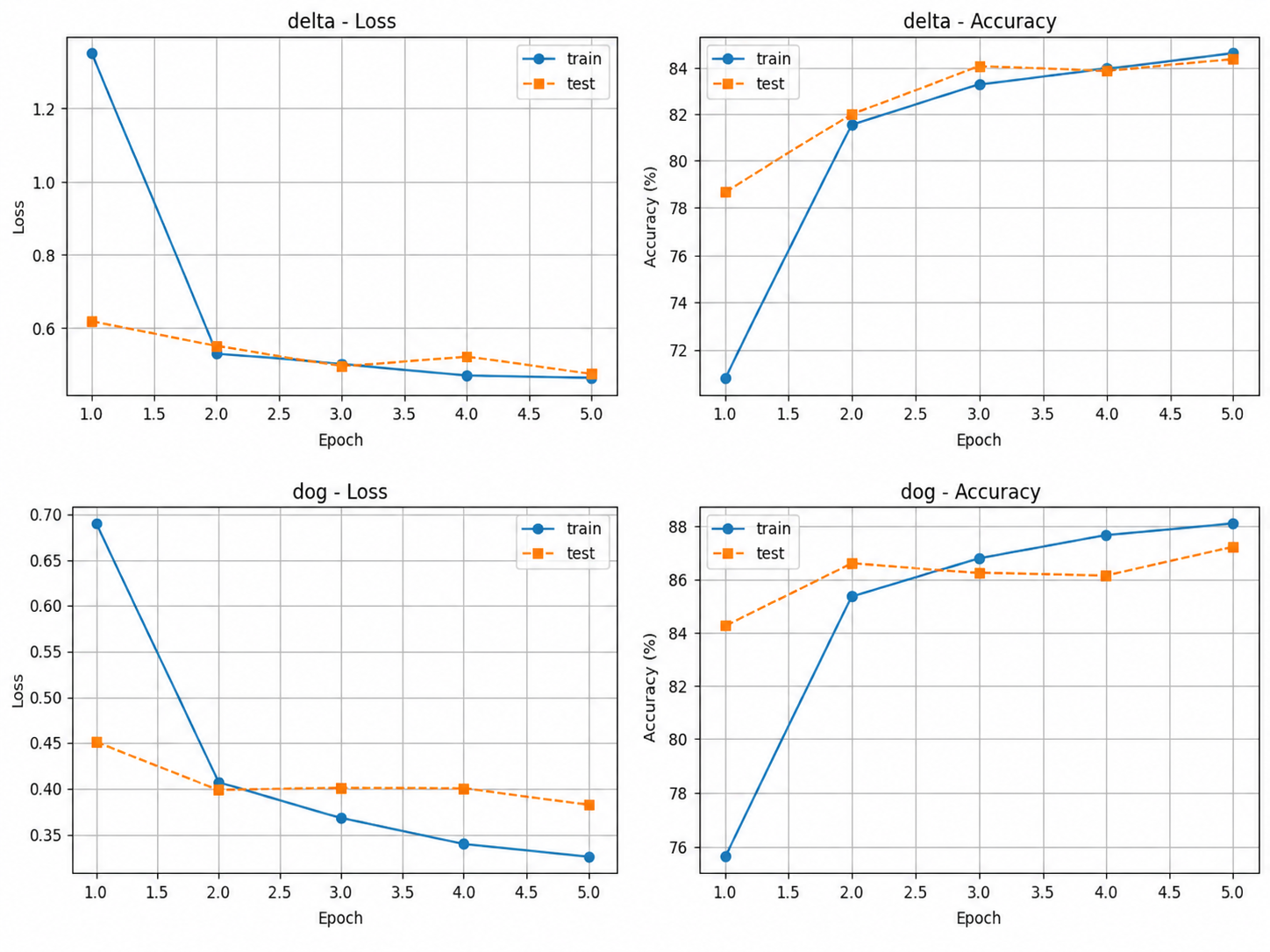}
    \caption{Training accuracy/loss vs. Testing accuracy/loss for delta and dog encoding (K-FAC)}
\end{figure}

K-FAC raised accuracy slightly over the Adam-trained SNN baseline; the improvement was consistent across runs but small in absolute terms, consistent with the expectation that second-order methods provide the most benefit on harder tasks where the loss landscape is less well-conditioned.

\subsubsection{Comparison to Non-Spiking Baselines} 

An important context for interpreting SNN accuracy numbers is the gap relative to non-spiking baselines of equivalent parameter count. 
On MNIST, a two-layer MLP with 256 and 128 hidden units achieves around 95\% accuracy, which the rate-coded SNN at high $T$ and $h$ matches closely.
On Fashion-MNIST, the same MLP architecture achieves approximately 89--90\%, while the SNN with rate coding achieves approximately 85--87\%, a gap of 3-4 percentage points that represents the cost of the spiking nonlinearity under first-order optimization. 

\subsubsection{Failure Modes and Limitations}

Several failure modes were observed across experimental configurations. 
Latency coding proved sensitive to the choice of time constant \(\tau\):
values well-matched to one dataset's intensity distribution caused
either silent neurons (when \(\tau\) was too small and the threshold was
never reached for dim pixels) or near-simultaneous firing (when \(\tau\)
was too large and all neurons fired within the first few time steps).
The Cholesky algorithm of K-FAC sometimes fails even when the damping mechanism 
is employed when dealing with very small batch sizes. In such cases, it is 
necessary to switch to using the pseudo-inverse, but the pseudo-inverse has 
poorer numerical stability. Rouser's threshold learning was observed to
be unstable in early training epochs when the learning rate ratio
\(\eta_\vartheta / \eta_w\) was set above approximately 5, suggesting
that the threshold adaptation dynamics are more sensitive to
initialization than the weight dynamics. These failure modes motivate
future work on adaptive damping for K-FAC in the SNN setting and
warm-start procedures for Rouser threshold initialization.

\section{Conclusions}
\subsection{Key Insights}
The central thesis of this work is that biologically-motivated input encoding can enable \textbf{smaller SNNs to match larger conventional networks} via efficient encoding and network optimization. 
Our experiment results support three concrete claims.
 
\textbf{Encoder choice is a parameter-efficiency lever, not an absolute-accuracy lever.} 
On Fashion-MNIST, a DoG-encoded SNN at $h{=}16$ (25k parameters) matches a delta-encoded SNN at $h{=}64$ (103k), which represents a 4$\times$ parameter saving at matched accuracy. 
At $h{=}32$ (51k), the DoG-encoded SNN exceeds the delta-encoded $h{=}256$ network (436k) by $+2.7$ percentage points while using
8.6$\times$ fewer parameters. 
This becomes advantageous in energy and capacity constrained scenarios such as neuromorphic hardware and biological systems.

\textbf{Adaptive thresholds (Rouser) provide a modest, consistent accuracy improvement at negligible additional cost.} 
On Fashion-MNIST, adaptive thresholds raise mean accuracy from 87.44\%
to 87.75\% across three seeds by adding one learnable parameter per neuron.
The gain is modest because Fashion-MNIST is a static-image dataset; we expect larger benefits on event-based or motion inputs, where per-neuron threshold adaptation can track the input distribution dynamically.

\textbf{K-FAC yields a consistent, modest improvement over Adam at this scale}
The improvement in accuracy achieved by K-FAC is in line with theoretical expectations: for simple datasets that are already close to saturation, the additional signal provided by the second-order curvature information is limited. 
It is expected that its benefits will be greater in more difficult tasks and in environments with complex loss surfaces. 
Rouser's adaptive threshold mechanism enables individual neurons to adjust their excitability, which is particularly beneficial when neural networks must process input with changing statistical characteristics. 
The dual learning rate scheme can accelerate the adjustment of the threshold without compromising the stability of weight learning.

Taken together, these three observations move SNNs toward the operational regime where their efficiency advantages become genuinely realisable on non-trivial image and vision tasks.

\subsection{Future Work}
Future work should evaluate our encoder-efficiency and optimization performance on more complex vision tasks. 
One such tasks would be content-rich complex image datasets such as like Oxford-IIIT Pet dataset, which would demand fine-feature extraction and warrant the detailed evaluatoin of convolutional SNNs.
Continued work can examine the temporal mechanisms behind delta-based encoders and Rouser's adaptive thresholds to exploit sparsity inherent of event-based datasets, potentially yielding significant gains in energy efficiency without compromising inference accuracy.

\bibliographystyle{IEEEtran}
\bibliography{citations}

\begin{thebibliography}{10}

\providecommand{\url}[1]{#1}
\csname url@samestyle\endcsname
\providecommand{\newblock}{\relax}
\providecommand{\bibinfo}[2]{#2}
\providecommand{\BIBentrySTDinterwordspacing}{\spaceskip=0pt\relax}
\providecommand{\BIBentryALTinterwordstretchfactor}{4}
\providecommand{\BIBentryALTinterwordspacing}{\spaceskip=\fontdimen2\font plus
\BIBentryALTinterwordstretchfactor\fontdimen3\font minus
  \fontdimen4\font\relax}
\providecommand{\BIBforeignlanguage}[2]{{%
\expandafter\ifx\csname l@#1\endcsname\relax
\typeout{** WARNING: IEEEtran.bst: No hyphenation pattern has been}%
\typeout{** loaded for the language `#1'. Using the pattern for}%
\typeout{** the default language instead.}%
\else
\language=\csname l@#1\endcsname
\fi
#2}}
\providecommand{\BIBdecl}{\relax}
\BIBdecl

\bibitem{lecun1998mnist}
Y.~LeCun, ``The {MNIST} database of handwritten digits,'' 1998.

\bibitem{xiao2017fashion}
H.~Xiao, K.~Rasul, and R.~Vollgraf, ``Fashion-mnist: a novel image dataset for
  benchmarking machine learning algorithms,'' \emph{arXiv preprint
  arXiv:1708.07747}, 2017.

\bibitem{Zenke2018}
F.~Zenke and S.~Ganguli, ``{SuperSpike}: Supervised learning in multi-layer
  spiking neural networks,'' \emph{Neural Computation}, vol.~30, no.~6, pp.
  1514--1541, 2018.

\bibitem{Neftci2019}
E.~O. Neftci, H.~Mostafa, and F.~Zenke, ``Surrogate gradient learning in spiking
  neural networks,'' \emph{IEEE Signal Processing Magazine}, vol.~36, no.~6, pp.
  51--63, 2019.

\bibitem{pfeiffer2018deep}
M.~Pfeiffer and T.~Pfeil, ``Deep learning with spiking neurons: Opportunities
  and challenges,'' \emph{Frontiers in Neuroscience}, vol.~12, p. 409662, 2018.

\bibitem{borst2023flies}
A.~Borst and L.~N. Groschner, ``How flies see motion,'' \emph{Annual Review of
  Neuroscience}, vol.~46, no.~1, pp. 17--37, 2023.

\bibitem{Rodieck1965}
R.~W. Rodieck, ``Quantitative analysis of cat retinal ganglion cell response to
  visual stimuli,'' \emph{Vision Research}, vol.~5, no.~12, pp. 583--601, 1965.

\bibitem{martinez2004role}
S.~Martinez-Conde, S.~L. Macknik, and D.~H. Hubel, ``The role of fixational eye
  movements in visual perception,'' \emph{Nature Reviews Neuroscience}, vol.~5,
  no.~3, pp. 229--240, 2004.

\bibitem{Davies2018}
M.~Davies \emph{et~al.}, ``Loihi: A neuromorphic manycore processor with on-chip
  learning,'' \emph{IEEE Micro}, vol.~38, no.~1, pp. 82--99, 2018.

\bibitem{Atick1992}
J.~J. Atick and A.~N. Redlich, ``What does the retina know about natural
  scenes?'' \emph{Neural Computation}, vol.~4, no.~2, pp. 196--210, 1992.

\bibitem{barlow1961possible}
H.~B. Barlow \emph{et~al.}, ``Possible principles underlying the transformation
  of sensory messages,'' \emph{Sensory Communication}, vol.~1, no.~01, pp.
  217--233, 1961.

\bibitem{atick1992could}
J.~J. Atick, ``Could information theory provide an ecological theory of sensory
  processing?'' \emph{Network: Computation in Neural Systems}, vol.~3, no.~2,
  pp. 213--251, 1992.

\bibitem{Gjorgjieva2014}
J.~Gjorgjieva, H.~Sompolinsky, and M.~Meister, ``Benefits of pathway splitting
  in sensory coding,'' \emph{Journal of Neuroscience}, vol.~34, no.~36, pp.
  12127--12144, 2014.

\bibitem{Takaghaj2025}
S.~M. Takaghaj and J.~Sampson, ``Rouser: Robust {SNN} training using adaptive
  threshold learning,'' in \emph{2025 IEEE International Conference on Artificial
  Intelligence Circuits and Systems (AICAS)}, 2025.

\bibitem{kingma2014adam}
D.~P. Kingma and J.~Ba, ``Adam: A method for stochastic optimization,''
  \emph{arXiv preprint arXiv:1412.6980}, 2014.

\bibitem{Martens2015}
J.~Martens and R.~Grosse, ``Optimizing neural networks with Kronecker-factored
  approximate curvature,'' in \emph{Proceedings of the 32nd International
  Conference on Machine Learning (ICML)}, 2015, pp. 2408--2417.

\bibitem{Guo2023}
W.~Guo, M.~E. Fouda, A.~M. Eltawil, and K.~N. Salama, ``Efficient training of
  spiking neural networks with temporally-truncated local backpropagation
  through time,'' \emph{Frontiers in Neuroscience}, vol.~17, p. 1047008, 2023.

\end{thebibliography}

\end{document}